\documentclass[dvipsnames]{article}

\usepackage[final]{openbmb_2025}

\usepackage{amsmath}
\usepackage{amssymb}
\usepackage{mathtools}
\usepackage{amsthm}
\usepackage{multirow}
\usepackage{makecell}
\usepackage{subcaption}
\usepackage{wrapfig}
\usepackage{graphicx}
\usepackage{pifont}

\usepackage[utf8]{inputenc} 
\usepackage[T1]{fontenc}    
\usepackage{url}            
\usepackage{booktabs}       
\usepackage{amsfonts}       
\usepackage{nicefrac}       
\usepackage{microtype}      

\usepackage{color, soul}
\usepackage[most,breakable]{tcolorbox}
\usepackage{xcolor}

\usepackage[labelfont=bf]{caption}
\usepackage{enumitem}
\usepackage{sectsty}
\usepackage{fontawesome5}

\usepackage{fancyhdr}
\usepackage{xcolor} 
\renewcommand{\headrulewidth}{1pt}
\makeatletter
\def\headrule{{\if@fancyplain\let\headrulewidth\plainheadrulewidth\fi
\hrule\@height\headrulewidth\@width\textwidth \vskip-\headrulewidth}}
\makeatother

\definecolor{BMBDarkBlue}{HTML}{315EFE}
\definecolor{BMBLightBlue}{HTML}{00D3ED}
\usepackage[colorlinks, linkcolor=BMBDarkBlue, anchorcolor=BMBDarkBlue, citecolor=BMBDarkBlue, urlcolor=BMBDarkBlue]{hyperref}

\sectionfont{\color{BMBDarkBlue}\fontfamily{zi4}\selectfont}
\subsectionfont{\color{BMBDarkBlue}\fontfamily{zi4}\selectfont}
\subsubsectionfont{\color{BMBDarkBlue}\fontfamily{zi4}\selectfont}

\newtcolorbox{mytheorem}{
  colback=gray!5,       
  colframe=gray!80,     
  boxrule=0.5pt,        
  arc=4pt,              
  left=4pt,             
  right=4pt,            
  top=4pt,              
  bottom=4pt,           
}

\usepackage{xspace}
\newcommand{\fancyheadname}{\textit{\textbf{AgentCPM-Report}}}
\setlist[itemize]{leftmargin=1em}
\setlist[enumerate]{leftmargin=1.5em}

\usepackage{setspace}
\usepackage[most]{tcolorbox} 

\usepackage[table]{xcolor}
\usepackage{makecell}
\definecolor{mygray}{RGB}{230,230,230}
\usepackage{fvextra}

\usepackage[most]{tcolorbox}
\tcbuselibrary{listings, breakable}

\title{AgentCPM-Report: Interleaving Drafting and Deepening \\ for Open-Ended Deep Research}

\author{%
\\
\bf AgentCPM Team
} 

\def\huggingface{\raisebox{-1.5pt}{\includegraphics[height=1.05em]{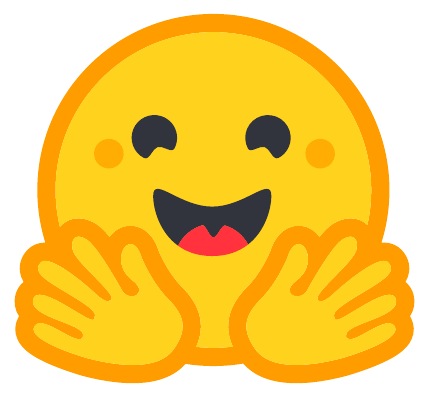}}}
\newcommand{\hflink}{https://huggingface.co/openbmb/AgentCPM-Report}

\newcommand{\ourmethod}[0]{WARP}

\begin{document}

\maketitle
\thispagestyle{fancy} 

\vspace{1em}
\begin{abstract}
Generating deep research reports requires large-scale information acquisition and the synthesis of insight-driven analysis, posing a significant challenge for current language models. Most existing approaches follow a \textit{plan-then-write} paradigm, whose performance heavily depends on the quality of the initial outline. However, constructing a comprehensive outline itself demands strong reasoning ability, causing current deep research systems to rely almost exclusively on closed-source or online large models. This reliance raises practical barriers to deployment and introduces safety and privacy concerns for user-authored data.
In this work, we present \textbf{AgentCPM-Report}, a lightweight yet high-performing local solution composed of a framework that mirrors the human writing process and an 8B-parameter deep research agent.
Our framework uses a \textbf{Writing As Reasoning Policy (WARP)}, which enables models to dynamically revise outlines during report generation. Under this policy, the agent alternates between \textbf{\textit{Evidence-Based Drafting}} and \textbf{\textit{Reasoning-Driven Deepening}}, jointly supporting information acquisition, knowledge refinement, and iterative outline evolution.
To effectively equip small models with this capability, we introduce a \textbf{Multi-Stage Agentic Training} strategy, consisting of cold-start, atomic skill RL, and holistic pipeline RL.
Experiments on DeepResearch Bench, DeepConsult, and DeepResearch Gym demonstrate that AgentCPM-Report outperforms leading closed-source systems, with substantial gains in \textit{Insight}. 
\end{abstract}

\section{Introduction}
\label{sec:intro}

Open-ended deep research requires artificial agents to navigate vast information landscapes and synthesize their findings into coherent, insightful reports~\citep{openaideepresearch, googledeepresearch, grok, perplexitydeepresearch, kimideepresearch, doubaodeepresearch}.
In the context of such complex inquiry, \textit{writing} is far more than the mere transcription of retrieved data. Instead, it reflects the \textit{knowledge-transforming} process described in cognitive psychology~\citep{scardamalia1987knowledge}: because the information landscape is initially opaque, researchers rarely execute a rigid, end-to-end plan derived solely from pre-existing thoughts. Rather, writing itself functions as a reasoning mechanism, progressively revealing what is not yet known.
Indeed, researchers often identify gaps, contradictions, or novel directions only during the act of drafting, indicating that effective synthesis depends on a tight and continual coupling between planning and writing.

\begin{figure}[t]
    \centering
    \includegraphics[width=0.95\linewidth]{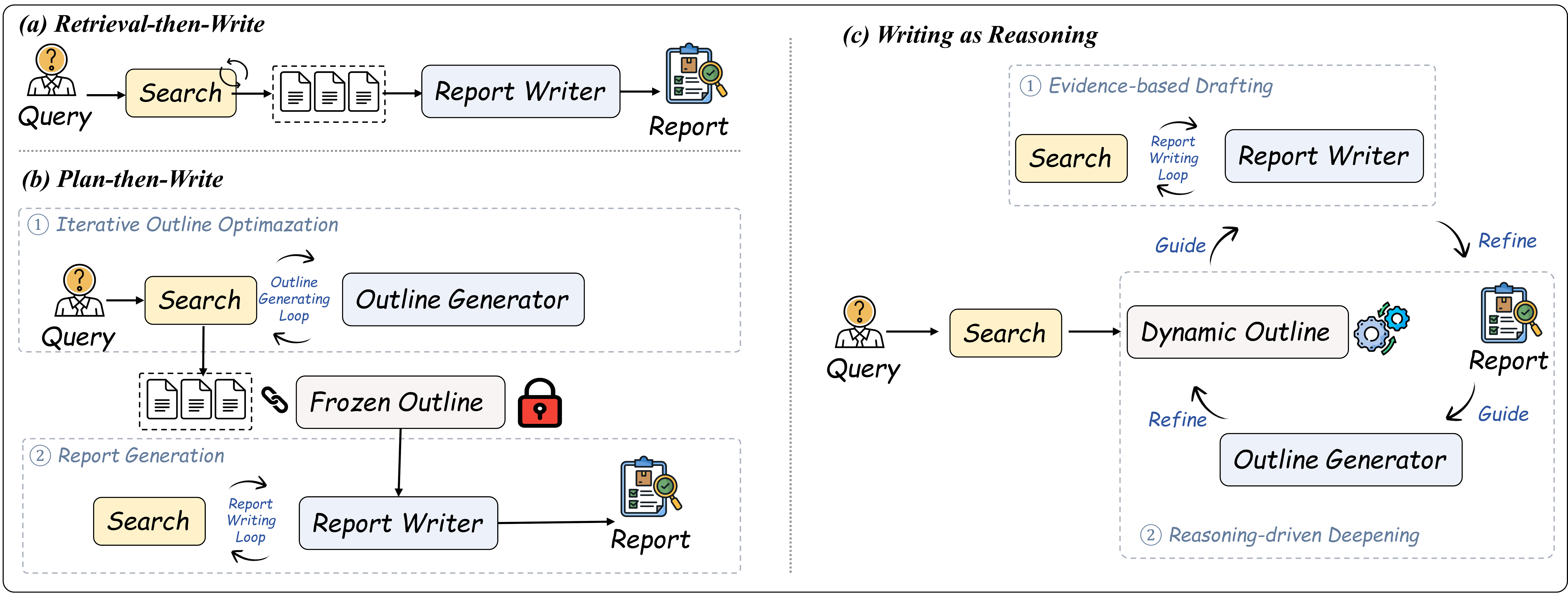}
    \caption{Comparison of different writing paradigms.}
    \label{fig: paradigm comparison}
\end{figure}

Despite this, existing deep research systems struggle to replicate such dynamics.
Early approaches followed a \textit{retrieval-then-write} paradigm (Fig.~\ref{fig: paradigm comparison}a)~\citep{hu2025step}, in which agents generated content sequentially based on retrieved evidence. While flexible, this loosely structured process frequently degenerates into incoherence over long horizons.
To improve structural consistency, more recent frameworks (Fig.~\ref{fig: paradigm comparison}b)~\citep{wang2024autosurvey, wang2025mapreduce-v2, yan2025surveyforge}, such as WebWeaver~\citep{li2025webweaver}, adopt a \textit{plan-then-write} paradigm. By freezing a comprehensive outline prior to writing, these systems enforce global structure and stability. However, this paradigm rests on the \textit{assumption of initial information completeness}—an assumption that is often violated in open-ended research.
By reducing the downstream writer to an executor of a static blueprint, this rigid separation prevents agents from capturing \textit{emergent insights}: subtle connections and refinements that surface only when articulating concrete arguments. As a result, such methods encounter an \textit{insight ceiling}, producing reports that are structurally sound yet intellectually shallow.

Another critical limitation of the \textit{plan-then-write} paradigm lies in its heavy reliance on generating a high-quality, comprehensive outline \emph{before} writing begins. This requirement places substantial demands on the model’s reasoning capacity and domain knowledge. Smaller models are generally weaker in these aspects than large-scale models, which has led most existing deep research systems to rely on closed-source or online large models as their backbone.
This reliance introduces a practical and often overlooked challenge: online deployment makes it difficult to support writing over users’ local or private data, as uploading such data inevitably raises security and privacy concerns. Consequently, there is a growing need for a fully local, on-device deep research and writing solution that does not depend on external large-scale models.

These two limitations stem from the same root cause: the rigid separation between planning and writing. To address both the \textit{insight ceiling} and the challenges of on-device deep research, we present \textbf{AgentCPM-Report}, a lightweight yet high-performance local system built upon a novel \textbf{WARP} (\textbf{W}riting \textbf{A}s \textbf{R}easoning \textbf{P}olicy) framework and an \textbf{8B-parameter} deep research agent.

WARP is a policy-level reformulation of deep research, motivated by the observation that any approach grounded in static planning inevitably incurs an insight ceiling. By modeling research as an iterative refinement loop, WARP enables planning decisions to emerge from, and adapt to, the writing process itself. Rather than adhering to a fixed outline, the agent alternates between two macro-states: \textit{Evidence-based Drafting} and \textit{Reasoning-driven Deepening}. Crucially, WARP is formulated as a dynamic policy instead of a rule-based heuristic. In the \textit{Reasoning-driven Deepening} state, the agent autonomously determines whether to terminate or continue deepening by evaluating the semantic density and logical coherence of the current draft. When further deepening is warranted, it decomposes high-level sections into more granular inquiries and updates the outline based on feedback from the writing process itself—closely mirroring the human knowledge-transforming process.

The dynamic nature of WARP introduces long-horizon credit assignment and a vastly expanded action space, which standard training pipelines fail to handle. 
We design a \textbf{Multi-Stage Agentic Training} strategy to ensure stable convergence under reasonable resource constraints.Specifically, we employ a \textit{trajectory pruning} mechanism to filter high-quality supervision signals, and design a curriculum-based reinforcement learning pipeline that progressively optimizes local atomic actions before fine-tuning end-to-end behavior. This training strategy yields a robust policy that adaptively balances research depth against computational cost, triggering recursive refinement only when it produces meaningful informational gains.

Extensive experiments on DeepResearch Bench, Deep Consult, and DeepResearch Gym demonstrate the effectiveness of AgentCPM-Report, yielding substantial improvements in overall report quality—particularly on the \textit{Insight} metric. Despite relying on only an 8B-parameter agent, AgentCPM-Report matches or outperforms the leading closed-source deep research systems, including \textit{Gemini-2.5-Pro}. These results indicate that a specialized WARP inference paradigm enables small-size, open-source models to achieve deep research capabilities previously associated with proprietary large-scale systems. Together, they establish a strong foundation for safe, privacy-preserving, and fully local deep research report generation.

\section{Method}
\label{sec:method}
In this section, we formally present AgentCPM-Report. First, we formulate deep research as a unified sequential decision-making process. Then, we provide our~\ourmethod\ (Writing As Reasoning Policy) framework. Finally, we show the multi-stage agentic training strategy.

\subsection{Problem Formulation}

Specifically, we formulate open-ended deep research as an iterative hierarchical decision-making process: 
At any interaction loop $i$, the agent observes a global state $S_i = (Q, O_i, D_i, C_i)$, comprising the user query $Q$, a dynamic outline $O_i$, the current draft $D_i$, and the context $C_i$ retrieved at the current loop $i$.
At the $j$-th step $t_{i,j}$ in a loop $i$, the agent executes an action $A_{i,j}$ selected from a defined action space: \{\textsc{Initialize, Search, Write, Expand, Terminate}\}.

This formulation unifies planning and writing: outline adjustments ($O_i \to O_{i+1}$) and content generation ($D_i \to D_{i+1}$) are treated as equivalent state transitions driven by the policy.

\begin{figure*}[t]
    \centering
    \includegraphics[width=0.95\linewidth]{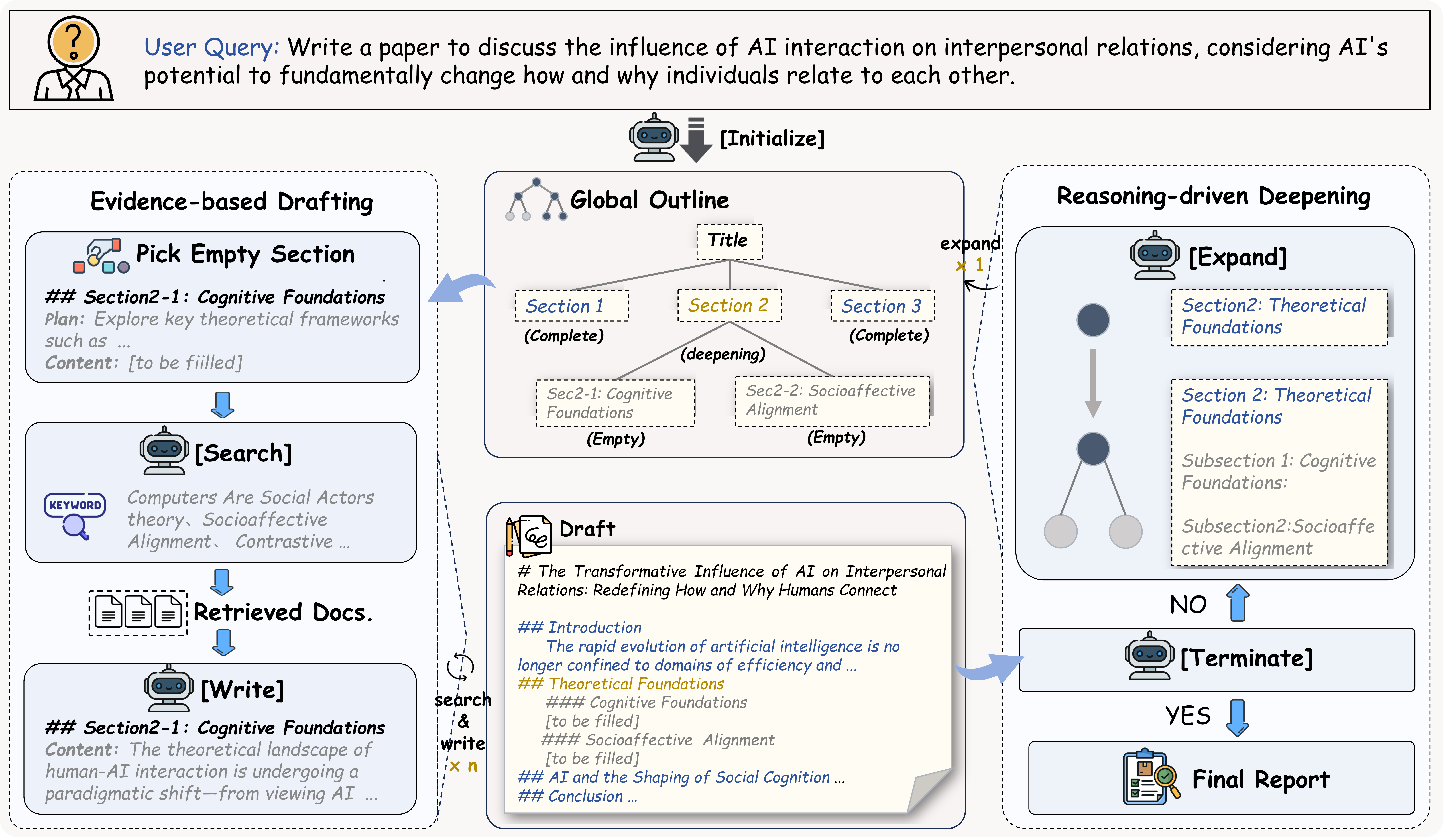}
    \caption{The \ourmethod\ framework. The agent interleaves \textbf{Evidence-Based Drafting} (writing content) and \textbf{Reasoning-Driven Deepening} (updating the 
    dynamic outline). This loop allows the agent to discover and bridge logical gaps that emerge only during the writing process.}
    \label{fig:warp_framework}
\end{figure*}

\subsection{The~\ourmethod\ Inference Diagram}

\ourmethod\ begins with a coarse-to-fine initialization strategy designed to establish a comprehensive research scope before diving into details. Starting from the initial state $S_0$, the agent analyzes the query $Q$ to generate broad search queries $q_0$. Upon retrieving the background context $C_0$, it synthesizes an initial Level-1 outline $O_0$:
\begin{equation}
    O_0 \leftarrow \textsc{Initialize} (Q, C_0).
\end{equation}
In contrast to static planners~\cite{li2025webweaver} that attempt to generate a fully detailed hierarchy, our $O_0$ is intentionally sparse, consisting only of high-level section titles and brief writing intents.
This design mitigates the risk of being ungrounded.

To maximize the final report quality, the agent operates under a unified policy $\pi_\theta$ that orchestrates the research trajectory. 
Crucially, this workflow is not linear but iterative, alternating between Drafting and Deepening.

\paragraph{Evidence-Based Drafting}
Given a tentative outline $O_i$, the agent executes a \textit{retrieve-then-write} strategy to convert the structural plan into substantiated content.
Unlike independent parallel generation, which often leads to fragmentation or redundancy, we enforce contextual consistency by conditioning retrieval queries on the accumulating narrative.
For a specific section $k$, the agent first formulates query $q_{i,k}$ based on user query $Q$, the section's intent $O_{i}^{k}$, and the draft context $D_{i}$:
\begin{equation}
    q_{i,k} \leftarrow \textsc{Search} (Q, O_i^k, D_i).
\end{equation}
Then, the retrieval tools will acquire new content $C_i^k$ based on the query $q_{i,k}$. This ensures that new information strictly extends the logical flow of previous sections.
The agent then synthesizes the section content $c_k$ by grounding the text in retrieved evidence to guarantee faithfulness:
\begin{equation}
    D_{i}^k \leftarrow D_{i}^{k-1} \oplus \textsc{Write}(Q, O_{i}^{k}, D_i^{k-1}, C_i^k).
\end{equation}
The objective is to achieve information integration—synthesizing disparate sources into a coherent argument—rather than mere aggregation.
This phase focuses on writing, iteratively populating the outline to produce a new draft $D_{i+1}$ that serves as the foundation for deeper reasoning.

\paragraph{Reasoning-Driven Deepening}
Initial outlines are inevitably constrained by the model’s pre-retrieval knowledge, which often creates an \emph{insight ceiling}: the structure may cover the breadth of the topic but fail to capture its nuanced depth.
To break this ceiling, the policy $\pi_\theta$ periodically shifts from local drafting to global planning, treating the newly generated draft $D_{i+1}$ as a fresh observation for reasoning and diagnosis.

Since $D_{i+1}$ provides a concrete reasoning context, the agent can detect logical gaps or superficial arguments that were invisible during initial planning. If section $k^*$ lacks depth, the agent generates a \textbf{local sub-sections} to decompose the topic, updating $O_{i}$ and triggering a targeted drafting cycle: 
\begin{equation}
    O_{i+1} \leftarrow O_{i} \oplus \textsc{Expand}\{k^*\}(Q, O_i, D_{i+1}).
\end{equation}
The process concludes only when the agent verifies that the logical chain is complete and the content depth aligns with the query's complexity:
\begin{equation}
    End \leftarrow \textsc{Terminate}(Q, O_i, D_{i+1}).
\end{equation}

\subsection{Multi-Stage Agentic Training}
\label{sec:training}

While large-scale models have demonstrated strong capabilities within our~\ourmethod\ framework (see~\S\ref{prompt-based LM}), training small-scale models (such as $8$B) for open-ended research is non-trivial. The challenges are twofold: (1) \textit{Ambiguous Termination}: even teacher models struggle to determine the optimal stopping point for research; (2) \textit{Sparse Rewards}: long horizons make reward assignment difficult.
To address these, we first introduce the trajectory pruning strategy in in \S\ref{sec:data_prep}. Then, we propose a curriculum learning pipeline in \S\ref{sec:rl_stages}.

\begin{figure*}[t]
    \centering
    \includegraphics[width=0.95\linewidth]{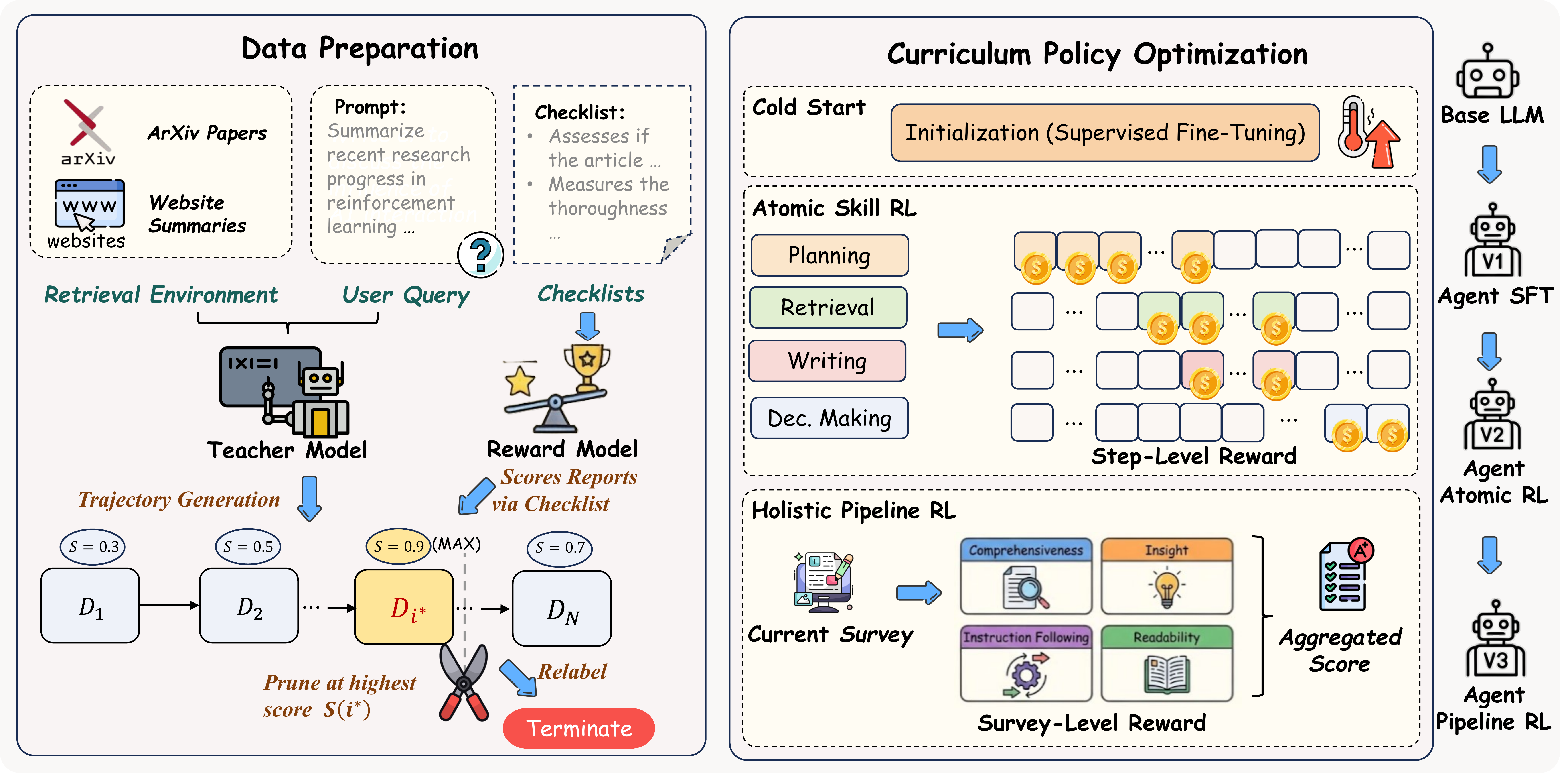}
    \caption{Overview of our multi-stage agentic training process.}
    \label{fig: training_process}
\end{figure*}

\begin{table*}[t]
  \small
  \centering
  \caption{Reward definition for \textbf{Atomic Skill RL}. For each metric, we indicate whether it requires references (\textbf{Ref.?}) or relies on the LLM-as-Judge method (\textbf{LLM?}).}
  \label{tab:atomic_rewards}
  \setlength\tabcolsep{1.6mm}{
  \begin{tabular}{llccl}
    \toprule
    \textbf{Ability} & \textbf{Metric Name} & \textbf{Ref.?} & \textbf{LLM?} & \textbf{Description \& Reward Objective} \\
    \toprule
    \multirow{3}{*}{\textbf{Planning}} 
      & Basic Properties & \ding{55} & \ding{55} & Section number and language consistency. \\
      & Holistic Quality & \ding{51} & \ding{51} & Quality metrics such as guidance, logic, clarity, etc. \\
      & Faithfulness & \ding{55} & \ding{51} & Whether the content in the plan is real and reliable. \\
    \midrule
    \textbf{Retrieval} 
      & Relevance Recall & \ding{51} & \ding{55} & Overlap score between retrieved docs and golden ones. \\
    \midrule
    \multirow{4}{*}{\textbf{Writing}} 
      & Basic Properties & \ding{55} & \ding{55} & Content length, citation number, and language consistency. \\
      & Holistic Quality & \ding{51} & \ding{51} & Quality metrics such as relevance, coverage, depth, etc.\\
      & Faithfulness & \ding{55} & \ding{51} & Penalizes unsupported claims. \\
      & Citation Precision & \ding{51} & \ding{55} &   Rewards citations that overlaps with golden ones. \\
    \midrule
    \textbf{Decision-Making}
      & Accuracy & \ding{51} & \ding{55} & Whether terminate at the suitable time. \\
    \bottomrule
  \end{tabular}}
  \label{tab:reward_functions}
\end{table*}

\subsubsection{Data Preparation}
\label{sec:data_prep}

A critical challenge in training is the scarcity of expert trajectories that exhibit efficient decision-making. Teacher models often either expand indefinitely or terminate arbitrarily, giving rise to what we term the \textbf{optimal stopping problem}.
To solve this, we introduce \textbf{trajectory pruning} strategy. Instead of cloning the teacher's termination behavior, we force the teacher to "over-expand" recursively. This generates a sequence of drafts with varying granularity $\{D_1, \dots, D_N\}$. We then retroactively identify the optimal point $i^*$ where the report draft $D_{i^*}$ has the highest score. We prune the trajectory at $i^*$, relabeling the action to \textsc{Terminate}. This provides a supervision signal for \textit{information saturation}, teaching the agent to stop based on report quality rather than arbitrary imitation.
Details on our query construction and retrieval environment are provided in App.~\ref{sec: user_query} and App.~\ref{sec: retrieval_env}.

\subsubsection{Curriculum Policy Optimization}
\label{sec:rl_stages}

Our optimization includes three stages, as shown in Figure~\ref{fig: training_process}: (1) \textbf{SFT for Cold Start:} Establishes basic instruction following and format adherence. (2) \textbf{Atomic Skill RL:} Uses teacher trajectories as anchors to master local execution and stabilize exploration. (3) \textbf{Holistic Pipeline RL:} Optimizes global report quality, enabling the agent to refine its strategy beyond the teacher's limitations.

\paragraph{Atomic Skill RL}

To tackle the reward assignment problem, we first decompose the global objective to atomic abilities: \textbf{planning} (\textit{Initialize}, \textit{Expand}), \textbf{retrieval} (\textit{Search}), \textbf{writing} (\textit{Write}), and \textbf{decision-making} (\textit{Terminate}). Then, we design different reward functions for them, which combine execution results (e.g., basic properties, holistic quality, and faithfulness) with \textit{reference alignment}, as shown in Table~\ref{tab:reward_functions}. This stage ensures the agent masters the "how"—producing valid plans, precise searches, and coherent paragraphs—before attempting to optimize global strategy.

\paragraph{Holistic Pipeline RL}
Local correctness (e.g., a valid paragraph) does not guarantee global coherence. Thus, the final stage shifts to end-to-end optimization to evaluate the final report quality, such as \textit{Comprehensiveness} and \textit{Insight}.
Crucially, this stage empowers the agent to deviate from the teacher's path. By propagating the holistic report score backward, the agent learns to trigger the \textit{deepening} only when it yields significant informational gain. This effectively refines the quality-efficiency frontier, suppressing redundant expansions that the teacher might have made.

\section{Experiments}

\subsection{Settings}

\paragraph{Implementation Details.}
We implement AgentCPM-Report using \textit{MiniCPM4.1-8B}~\citep{team2025minicpm4} as the backbone for our deep research agent system. Training follows the curriculum described in \S\ref{sec:training}, progressively scaling from atomic skills to holistic reporting.
During the training and inference phase, we cap the report structure at three levels and limit the number of deepening steps to $12$ to ensure efficiency. Detailed hyperparameters and data statistics per stage are provided in App.~\ref{app: training_details}.

\paragraph{Benchmarks and Metrics.}
To ensure comprehensive evaluation, we test on three diverse benchmarks:
(1) \textbf{DeepResearch Bench}~\citep{du2025deepresearchbench} (100 PhD-level scientific tasks);
(2) \textbf{DeepConsult}~\citep{DeepConsult} (102 business and financial analysis queries); and
(3) \textbf{DeepResearchGym}~\citep{coelho2025deepresearchgym} (100 general-purpose information-seeking tasks).
We adhere to the standard evaluation protocols of each benchmark, employing \textit{Gemini-2.5-Pro}, \textit{o3-mini}, and \textit{GPT-4.1-mini} respectively as impartial judges.

\paragraph{Baselines.}
We compare AgentCPM-Report against three distinct categories of latest systems: (1) \textbf{Proprietary Systems:} Leading commercial deep research systems including OpenAI~\citep{openaideepresearch}, Gemini~\citep{googledeepresearch}, and Claude~\citep{anthropic}, and Doubao~\citep{doubaodeepresearch}.
(2) \textbf{Prompt-Based Frameworks:} WebWeaver~\citep{li2025webweaver}, Enterprise DR~\citep{prabhakar2025enterprise}, and RhinoInsigh~\citep{lei2025rhinoinsight}.
(3) \textbf{Trained Open Models:} Recent open-source research agents including WebShaper~\citep{tao2025webshaper}, WebThinker~\citep{li2025webthinker} and DR Tulu~\citep{shao2025drtulu}.

\subsection{Main Results}

\begin{figure*}[htbp]
    \centering
    \includegraphics[width=\linewidth]{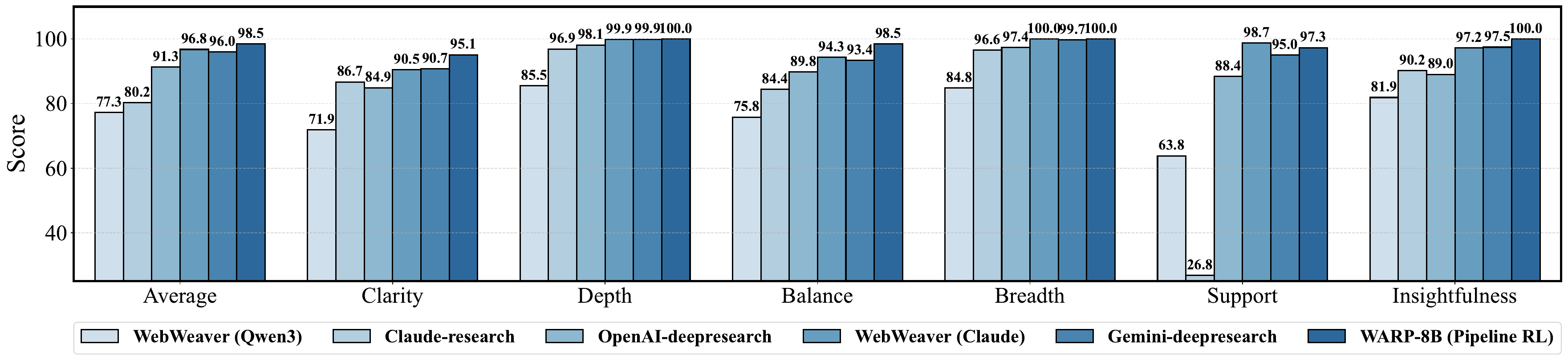}
    \caption{Performance of agent systems on DeepResearch Gym.}
    \label{fig: deepresearchgym}
\end{figure*}

Our results on three benchmarks are summarized in Table~\ref{tab:deepresearch_bench} and Figure~\ref{fig: deepresearchgym}.

\paragraph{(1) Our WARP framework has strong performance on \textit{Insight} and \textit{Comprehensiveness}.}

Across these benchmarks, our method achieves nearly the best performance in both \textit{Insight} and \textit{Comprehensiveness} metrics despite using the smallest model. Specifically, on the DeepResearch Bench, it achieves an \textit{Insight} score of $52.64$ and a \textit{Comprehensiveness} score of $50.54$, surpassing Gemini-2.5-Pro-deepresearch ($49.45$ and $49.51$, respectively). On the DeepResearch Gym, it gets the highest $100.0$ score in the \textit{Depth}, \textit{Breadth}, and \textit{Insightfulness} metrics.
These gains stem directly from our \emph{reasoning-driven deepening}. On one hand, the agent continuously \textbf{extracts insights from condensed intermediate drafts}, enabling deeper reasoning and synthesis. On the other hand, by revisiting intermediate outputs, it can \textbf{identify missing topics and globally assess which sections require further expansion}, resulting in broader and more balanced coverage.

\begin{table*}[htbp]
    \centering
    \caption{Performance of agent systems on DeepResearch Bench in terms of comprehensiveness (Comp.), insight, instruction-following (Inst.), readability (Read.) and DeepConsult (Avg., Win, Tie, Lose).}
    \setlength{\tabcolsep}{2mm}{
    \small
    \begin{tabular}{lccccccccc}
    \toprule
    \multirow{2}{*}{\textbf{Agent systems}} & \multicolumn{5}{c}{\textbf{DeepResearch Bench}} & \multicolumn{4}{c}{\textbf{DeepConsult}} \\
    \cmidrule(lr){2-6} \cmidrule(lr){7-10}
     & Overall & Comp. & Insight & Inst. & Read. & Avg. & Win & Tie & Lose \\
    \midrule
    \multicolumn{10}{l}{\textit{\textbf{Proprietary Deep Research Systems}}} \\
    Doubao-research & 44.34 & 44.84 & 40.56 & 47.95 & 44.69 & 5.42 & 29.95 & 40.35 & 29.70 \\
    Claude-research & 45.00 & 45.34 & 42.79 & 47.58 & 44.66 & 4.60 & 25.00 & 38.89 & 36.11 \\
    OpenAI-deepresearch & 46.45 & 46.46 & 43.73 & 49.39 & 47.22 & 5.00 & 0.00 & 100.00 & 0.00 \\
    Gemini-2.5-Pro-deepresearch & \textbf{49.71} & \textbf{49.51} & \textbf{49.45} & \textbf{50.12}  & \textbf{50.00} & \textbf{6.70} & 61.27 & 31.13 & \textbf{7.60} \\

    \midrule
    \multicolumn{10}{l}{\textit{\textbf{Prompt-Based Frameworks}}} \\
    WebWeaver (Qwen3-30B-A3B) & 46.77 & 45.15 & 45.78 & 49.21 & 47.34 & 4.57 & 28.65 & 34.90 & 36.46 \\
    WebWeaver (Claude-Sonnet-4) & 50.58 & \textbf{51.45} & 50.02 & 50.81 & 49.79 & \textbf{6.96} & 66.86 & 10.47 & 22.67 \\
    Enterprise DR (Gemini-2.5-Pro) & 49.86 & 49.01 & 50.28 & 50.03 & 49.98 & 6.82 & \textbf{71.57} & 19.12 & \textbf{9.31} \\
    RhinoInsigh (Gemini-2.5-Pro) & \textbf{50.92} & 50.51 & \textbf{51.45} & \textbf{51.72} & \textbf{50.00} & 6.82 & 68.51 & 11.02 & 20.47 \\

    \midrule
    \multicolumn{10}{l}{\textit{\textbf{Trained Open Models}}} \\
    WebShaper-32B & 34.93 & 31.58 & 26.17 & 44.81 & 40.38 & 1.63 & 3.25 & 3.75 & 93.00 \\
    WebThinker-32B-DPO  & -- & 39.40 & 35.40 & 46.00 & \textbf{43.50} & -- & -- & -- & -- \\
    DR Tulu-8B     & -- & \textbf{41.70} & \textbf{41.80} & \textbf{48.20} & 41.30 & -- & -- & -- & -- \\

    \midrule
    \multicolumn{10}{l}{\textit{\textbf{Our Deep Research Systems}}} \\
    AgentCPM-Report (SFT)      & 46.73 & 46.24 & 48.10 & 47.61 & 41.79 & 6.04 & 54.17 & 10.29 & 35.54 \\
    AgentCPM-Report (Atomic RL)  & 48.81 & 48.70 & 51.36 & 48.64 & 42.25 & 6.06 & 56.13 & 11.03 & 32.84 \\
    AgentCPM-Report (Pipeline RL)  & \textbf{50.11} & \textbf{50.54} & \textbf{52.64} & \textbf{48.87} & \textbf{44.17} &\textbf{6.60} & \textbf{57.60} & 13.73 & \textbf{28.68} \\
    \bottomrule
    \end{tabular}}
    \label{tab:deepresearch_bench}
\end{table*}

\paragraph{(2) The \textit{Multi-Stage Agentic Training} brings stable and comprehensive improvement.}

The performance of AgentCPM-Report steadily improves from SFT to Atomic RL and finally to Pipeline RL across all metrics on these benchmarks. On DeepResearch Bench, the metric \textit{comprehensiveness} rises from $46.24$ to $50.54$, \textit{Insight} from $48.10$ to $52.64$, and \textit{Readability} from $41.79$ to $44.17$. On DeepConsult, average score grows from $6.04$ to $6.60$, win rate increase from $54.17$\% to $57.60$\%, and loss rate drop from $35.54$\% to $28.68$\%. These consistent gains demonstrate that each stage of the curriculum contributes to mastering the full deep research workflow, yielding a more stable and capable agent.

\paragraph{(3) Small-scale agent systems can rival large-scale ones.}

Averaged across benchmarks, our deep research system demonstrates excellent performance. Our AgentCPM-Report (Pipeline RL) achieves an \textit{Overall} score of $50.11$ on DeepResearch Bench, surpassing Gemini-2.5-Pro-deepresearch ($49.71$). It also attains state-of-the-art results on DeepResearch Gym, with an average score of $98.48$.
These results show that integrating WARP with \textit{multi-stage agentic training} enables small models to reach the performance level of leading proprietary research systems. These findings suggest that, for deep research tasks, the primary bottleneck lies not in model size, but in the design of effective cognitive and planning processes that fully leverage a model’s inherent capabilities.

\subsection{Analysis}

\subsubsection{Does WARP remain effective without training?}
\label{prompt-based LM}

To assess whether our framework is intrinsically effective without training, we conduct a prompt-based comparison on DeepResearch Bench using a larger model, \textit{Qwen3-235B-A22B-Instruct-2507}~\citep{yang2025qwen3}. We compare two policies: (1) \textit{Plan-then-write}, where the model first constructs a detailed outline through retrieval and then generates the report from this fixed plan; and (2) WARP, which starts from a simple outline and interleaves writing with iterative deepening.

\begin{table}[htbp]
    \centering
    \small
    \caption{Evaluation for different generation paradigms.}
    \setlength\tabcolsep{1.4mm}{
    \begin{tabular}{lccccc}
        \toprule
        \textbf{Paradigm} & \textbf{Overall} & \textbf{Comp.} & \textbf{Insight} & \textbf{Inst.} & \textbf{Read.} \\
        \midrule
        Plan-then-write & 49.90 & 49.35 & 51.60 & 50.13 & 46.46 \\ 
        WARP & \textbf{50.72} & \textbf{50.33} & \textbf{52.79} & \textbf{50.32} & \textbf{47.20} \\ 
        \bottomrule
    \end{tabular}}
    \label{tab:framework_selection}
\end{table}

As shown in Table~\ref{tab:framework_selection}, WARP consistently outperforms the \textit{Plan-then-write} paradigm across all metrics, with a notable gain in \textit{Insight} (+1.19) and \textit{Comprehensiveness} (+0.98). By using the evolving draft as a reasoning context, WARP can detect underdeveloped or ambiguous content during writing and trigger targeted \textit{Deepening} with additional evidence, whereas \textit{Plan-then-write} remains constrained by a static outline. This confirms that draft-aware deepening is a key driver of insight.

\subsubsection{How multi-stage training shapes agent actions and report structure?} 

In this section, we analyze how agent behavior evolves across training stages by examining statistics of its actions and report sections. Specifically, we focus on the \emph{Write} and \emph{Expand} actions, which directly determine the report structure.

\begin{table}[htbp]
    \centering
    \small
    \caption{Evolution of action usage and hierarchical sectioning across training stages on DeepResearch Bench.}
    \label{tab:action_stats}
    \setlength\tabcolsep{1.4mm}{
    \begin{tabular}{lccccc}
    \toprule
    \multirow{2}{*}{\textbf{Stages}} & \multicolumn{2}{c}{\textbf{Actions}} & \multicolumn{3}{c}{\textbf{Sections}} \\  
    \cmidrule(lr){2-3} \cmidrule(lr){4-6}
     & \textbf{Write} & \textbf{Expand} & \textbf{Level-1} & \textbf{Level-2} & \textbf{Level-3} \\
    \midrule
    SFT & 21.24 & 4.44 & 6.27 & 10.11 & 4.86 \\
    Atomic RL & 36.89 & 8.88 & 6.49 & 14.17 & 16.50 \\
    Pipeline RL & 39.51 & 8.63 & 6.52 & 15.75 & 17.32 \\
    \bottomrule
    \end{tabular}}
\end{table}

\begin{figure}[htbp]
    \centering
    \includegraphics[width=0.65\linewidth]{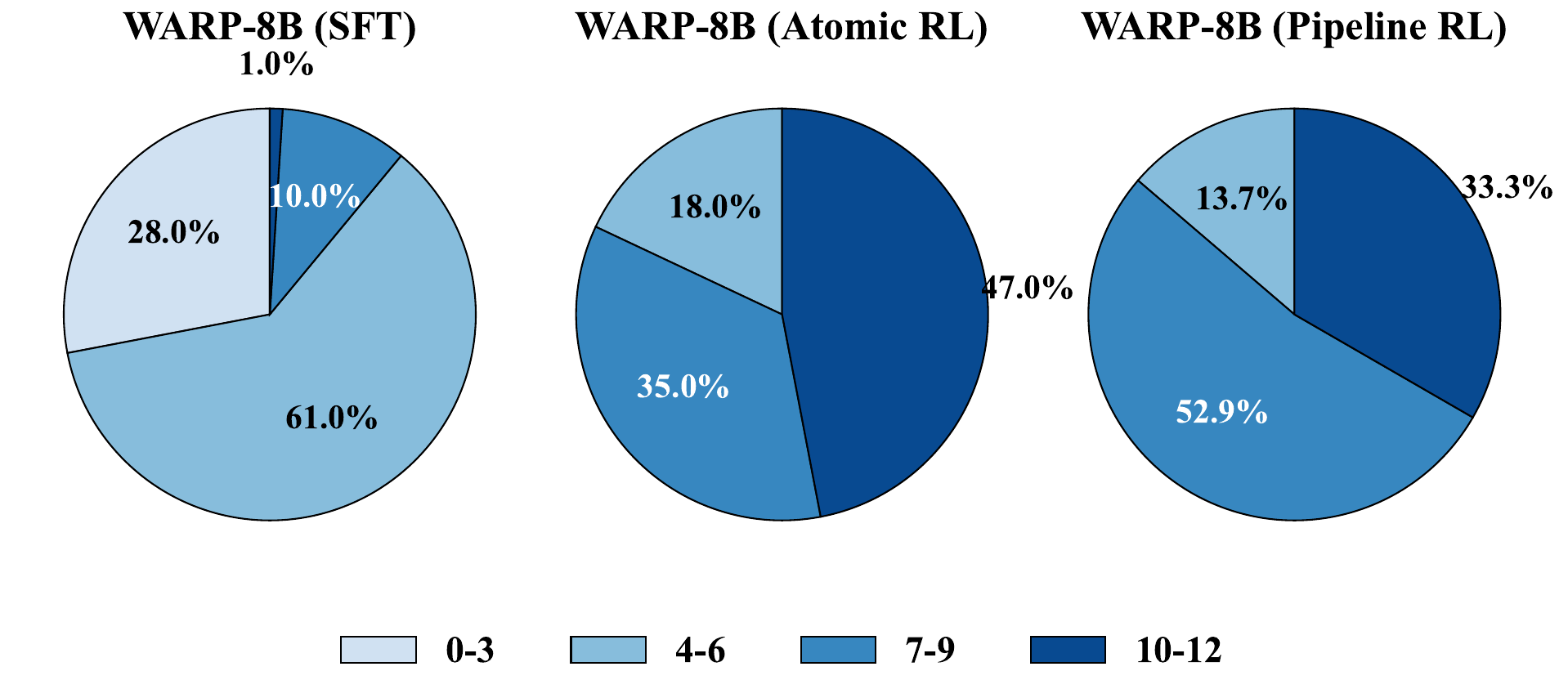}
    \caption{The \textit{Expand} steps on DeepResearch Bench.}
    \label{fig: deepen_pie}
\end{figure}

As shown in Table~\ref{tab:action_stats}, a clear behavioral shift emerges when moving from SFT to the RL-based training stages. The frequency of \textbf{\emph{Expand}} (Deepening) actions nearly doubles (from 4.44 to around 8.8), which in turn leads to a dramatic growth in fine-grained subsectioning (Level-3: from $4.86$ to $17.32$). Moreover, Figure~\ref{fig: deepen_pie} illustrates that RL training drives the agent to deepen more compared to SFT, ensuring at least 4 \textit{Expand} steps in all cases. This trend indicates that RL training effectively equips the agent with \textit{Reasoning-Driven Deepening}: Rather than adhering to a shallow outline as in SFT, the agent learns to identify underdeveloped parts of a draft and proactively expand them through iterative refinement, resulting in reports with substantially richer structure.

\subsubsection{How does the number of deepening influence report quality?} 

To examine whether the model has learned an appropriate stopping policy for \textit{deepening} and to quantify how deepening depth affects report quality, we conduct a "Forced Expansion" experiment.
Specifically, during inference, we force AgentCPM-Report (Pipeline RL) to apply the \emph{Expand} action exactly $k$ times, where $k$ ranges from 0 to 15.
We then compare this forced expansion curve with the actual deepening behaviors of AgentCPM-Report at different training stages (SFT, Atomic RL, and Pipeline RL), overriding their learned termination policies.
This allows us to directly evaluate report quality as a function of deepening depth and to compare the model’s learned stopping behavior against the optimal expansion point.

\begin{figure*}[htbp]
    \centering
    \includegraphics[width=\linewidth]{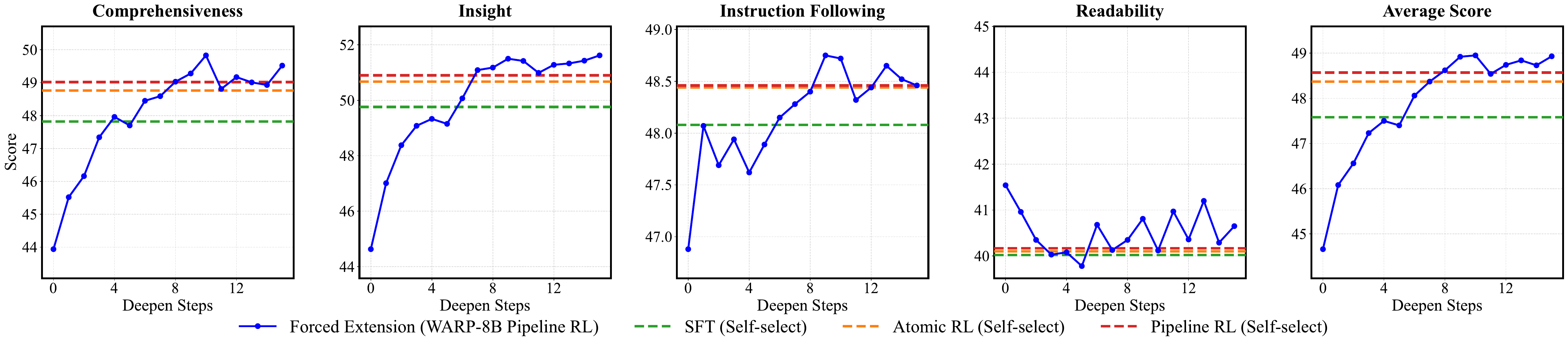}
    \caption{Mean Performance metrics per Deepen Steps on DeepResearch Bench.}
    \label{fig: deepen_pre_step}
\end{figure*}

The results in Figure~\ref{fig: deepen_pre_step} reveal three consistent patterns.
\textbf{First}, performance increases steadily with deeper expansion and begins to plateau at around nine steps, indicating diminishing returns beyond this depth.
\textbf{Second}, both \textit{Comprehensiveness} and \textit{Insight} rise strongly with deepening, improving by nearly $6$ points from shallow to sufficiently deep regimes, confirming the importance of iterative refinement for rich and insightful reports.
\textbf{Third}, different training stages exhibit distinct stopping behaviors. The SFT agent typically stops within $6$ steps and rarely reaches the saturation regime, whereas the Atomic RL and Pipeline RL agents shift their stopping distributions toward $6$–$15$ steps, closely matching the empirically optimal depth.

\subsubsection{How does the trajectory pruning affect the agent training?}

To address the optimal stopping problem, we introduce a \textit{trajectory pruning} strategy to construct higher-quality training data. In this section, we isolate its impact by training SFT models using the same teacher-generated trajectories, either with or without pruning. We consider two settings:
(1) w/o pruning, which directly uses the raw trajectories produced by the teacher model, and
(2) with pruning, which selects the best intermediate draft in one trajectory based on reward scores, retaining only the sub-trajectory before that draft.

\begin{table}[htbp]
    \centering
    \small
    \caption{Effect of trajectory pruning on SFT training.}
    \setlength\tabcolsep{1.2mm}{
    \begin{tabular}{lccccc}
        \toprule
        \textbf{Trajectory} & \textbf{Overall} & \textbf{Comp.} & \textbf{Insight} & \textbf{Inst.} & \textbf{Read.} \\
        \midrule
        w/o pruning  & 45.80 & 44.95 & 47.35 & 46.71 & 40.86 \\
        with pruning & \textbf{46.73} & \textbf{46.24} & \textbf{48.10} & \textbf{47.61} & \textbf{41.79} \\
        \bottomrule
    \end{tabular}}
    \label{tab:data_selection}
\end{table}

As shown in Table~\ref{tab:data_selection}, models trained on pruned trajectories consistently outperform those trained on raw teacher ones across all evaluation dimensions. This indicates that trajectory pruning effectively improves the quality of supervision.
More importantly, these results reveal a key limitation of large teacher models: although they generate strong drafts, their termination decisions are often suboptimal. By applying reward-based selection over intermediate states, trajectory pruning filters out poorly timed stopping points and provides cleaner training signals. As a result, the student model learns a more accurate termination policy, which is crucial for effective \textit{Reasoning-Driven Deepening}.



\section{Conclusion and Future Works}
In this work, we address the limitations of existing deep research systems that rigidly separate planning from writing, an assumption that leads to an inherent insight ceiling and drives reliance on large, closed-source models. We propose \textbf{AgentCPM-Report}, a fully local deep research system built upon \textbf{WARP} (Writing As Reasoning Policy), which reformulates deep research as a policy-level iterative refinement process where planning decisions dynamically emerge from the writing itself. To support the long-horizon decision-making and expanded action space induced by WARP, we introduce a multi-stage agentic training strategy that enables stable and efficient learning. Extensive experiments on multiple deep research benchmarks demonstrate substantial improvements in report quality—particularly in insight—showing that, despite relying on only an 8B-parameter model, AgentCPM-Report surpasses several closed-source systems. Overall, this work establishes a strong foundation for safe, privacy-preserving, and fully local deep research report generation and highlights policy design as a viable alternative to model scaling for advancing deep research capabilities.

\paragraph{Better report presentation.}
In most existing deep research systems, including ours, tables and figures are generated inline with paragraph-level text. However, constructing tabular layouts requires a reasoning process fundamentally different from writing prose, placing heavy demands on a model’s structural and formatting abilities. This coupling partly explains why agent systems based on smaller models often underperform large ones in presentation quality. A promising direction is to decouple presentation from content generation and assign it to a dedicated rendering agent, which could enable small models to achieve comparable or even superior layout quality. Moreover, current readability evaluation remains largely text-based and weakly reflects the true visual structure of rendered reports, suggesting the need for visual-modality evaluation in future work.

\paragraph{More information sources.}
Our system relies on a locally deployed textual knowledge base (e.g., arXiv abstracts and web summaries), which ensures stability and reproducibility but limits coverage and timeliness. It also lacks access to images, videos, domain-specific corpora, and personalized data. Future extensions will expand the knowledge base to support multi-modal content, local and personalized sources, and continuous updates, enabling richer and more realistic research scenarios.

\section{Contributions and Acknowledgments}
AgentCPM-Report is the result of the collective efforts of all members of our team.

\textbf{Project Leads:} Yishan Li, Wentong Chen

\textbf{Contributors:} Yishan Li, Wentong Chen, Yukun Yan, Mingwei Li, Sen Mei, Xiaorong Wang, Kunpeng Liu, Xin Cong, Shuo Wang, Zhong Zhang, Yaxi Lu, Zhenghao Liu, Yankai Lin, Zhiyuan Liu, Maosong Sun

\textbf{Advisors:} Yukun Yan, Yankai Lin, Zhiyuan Liu, Maosong Sun

\newpage

\bibliographystyle{citation}
\bibliography{citation}

\newpage

\appendix
\section{Method Details}
\label{sec:appendix A}

\subsection{The Prompts in \ourmethod\ Framework}
\label{sec: prompts}

In our \ourmethod\ framework, there are five actions in all three stages: \textit{Initialize}, \textit{Search}, \textit{Write}, \textit{Expand(Deepen)}, and \textit{Terminate}. In the \textbf{\textit{Initialization}} stage, the agent generates the initial Level-1 outline with writing plans by the \textit{search} and \textit{initialize} actions. The prompt for the \textbf{\textit{search}} is shown in Figure~\ref{fig: search}, and the prompt for the \textbf{\textit{initialize}} is shown in Figure~\ref{fig: initialize}. Then, in the \textbf{\textit{Evidence-Based Drafting}} stage, the agent writes the paragraphs by the \textit{search} and \textit{write} actions. The prompt for the \textbf{\textit{write}} is shown in Figure~\ref{fig: write}. After that, in the \textbf{\textit{Reasoning-Driven Deepening}} stage, the agent will decision whether to expand a section for more details by the \textbf{\textit{expand}} action or to end the total process directly by the \textbf{\textit{terminate}} action. The prompt for both actions is shown in Figure~\ref{fig: expand-terminate}.

\begin{tcolorbox}[
    breakable, 
    title=Prompt for Search Keywords,
    colback=gray!5!white, 
    colframe=gray!75!black,
    fontupper=\normalsize,
]
\begin{Verbatim}[breaklines=true]
You are a professional report generation expert, skilled at creating high-quality report outlines.  
Now, you need to analyze the user’s question and provide a simple article outline structure (only top-level sections).

** User Query **  
[user query]

** Latest Retrieved Information **  
[current information]

## Notes  
1. The outline must be comprehensive, logically sound, and aligned with the user’s stated preferences and requirements.  
2. The output language must match the language of the user’s query.

** Available Actions **  
- initialize: Generate the top-level section outline along with an appropriate title.

## Action Format:  
<action> {"name": "initialize", "title": "...", "sections": [{"title": "...", "plan": "..."}, {"title": "...", "plan": "..."}, ...]} </action>

** Output Format **  
<thought> Provide a detailed reasoning process </thought>  
<action> Action (in JSON format) </action>

Please output strictly according to the specified format.
\end{Verbatim}
\end{tcolorbox}
\captionof{figure}{The LLM prompt for the \textit{initialize} action.}\label{fig: initialize}

\begin{tcolorbox}[
    breakable, 
    title=Prompt for Search Keywords,
    colback=gray!5!white, 
    colframe=gray!75!black,
    fontupper=\normalsize,
]
\begin{Verbatim}[breaklines=true]
You are a searcher within a multi-agent system consisting of "Analyst-Searcher-Writer". You must perform retrieval based on instructions from the "Analyst". Carefully select the most accurate search keywords and strictly adhere to the specified output format.
You should focus on the user's query and the current article outline to determine the most relevant keywords for searching. You can give one or less to five keywords. The content should be in the same language as the user's query.

** User Query **
[user query]

** Current Article Outline **
[current outline]

** Analyst's Instruction **
[current instruction]

## Action Example:
<action> {"name": "search", "keywords": [keyword-1, keyword-2, ..]} </action>

** Output Format **
<thought> Your reasoning process </thought>
<action> Action (in JSON format) </action>

Please output strictly according to the specified format.
\end{Verbatim}
\end{tcolorbox}
\captionof{figure}{The LLM prompt for the \textit{search} action.}\label{fig: search}

\begin{tcolorbox}[
    breakable, 
    title=Prompt for Search Keywords,
    colback=gray!5!white, 
    colframe=gray!75!black,
    fontupper=\normalsize,
]
\begin{Verbatim}[breaklines=true]
You are a writer operating within a multi-agent system consisting of "Analyzer-Searcher-Writer". Based on instructions from the "Analyzer", the current writing status, and the most recently retrieved information, you are to compose a new paragraph while ensuring logical coherence and accurate citation of facts.
You should give a paragraph with breadth and depth, ensuring it is informative and engaging. You are encouraged to incorporate examples, **tables**, code snippets, and other elements to enhance the content. But don't write other sections or chapters that are not assigned to you.
You'd better give analytical and comparative content, not just a summary of facts. Please attention the coherence and logical flow of the entire article and the other sections. 
You can extract the claims from the retrieved information, and design how to write the paragraph based on the claims in the thought process.
**BE FAITHFUL! Make sure all your claims, especially the facts and the numbers, can be supported by the retrieved information, with your citations. Don't add any claims can't be supported by your citations. All the facts or data in your claims should can be found in the retrieved information you cited.**
**You should ensure that the content you write is not redundant with other sections.**
**And you should strictly follow the citation format like  \\cite{bibkey} or \\cite{bibkey1, bibkey2..} for any referenced information. The content should be in the same language as the user's query.**

PLEASE JUST OUTPUT THE CONTENT IN ANALYZER'S INSTRUCTION, DO NOT OUTPUT OTHER SECTIONS.

THE OUTPUT SHOULD BE IN THE SAME LANGUAGE AS THE USER'S QUERY.


** User Query **
[user query]

** Current Article Summary **
[current survey]

** Analyzer’s Instruction **
[current instruction]

** Retrieved Information **
[current information]

## Action Example:
<action> content </action>

** Output Format **
<thought> Your thought process </thought>
<action> Your Content (in Markdown format) (include BIBKEY for citations within the content) </action>

Please strictly follow the specified output format.
\end{Verbatim}
\end{tcolorbox}
\captionof{figure}{The LLM prompt for the \textit{write} action.}\label{fig: write}

\begin{tcolorbox}[
    breakable, 
    title=Prompt for Search Keywords,
    colback=gray!5!white, 
    colframe=gray!75!black,
    fontupper=\normalsize,
]
\begin{Verbatim}[breaklines=true]
You are a professional report-generation expert skilled at crafting high-quality report outlines.  
Based on the the user’s stated preferences, you must now determine whether any section requires expansion into subsections.

## Important Notes:
1. Select only the single section or subsection most in need of expansion.
2. If no expansion is needed, output a "terminate" (no operation) action.
3. If you think the expendsion is necessary to make the article more comprehensive or insightful, feel free to expand it.
4. Make sure the new subsections aren't redundant or overly detailed with other sections. If it's too detailed or redundant with other sections, just terminate it. 
5. Make sure the new subsections are relevant and coherent with other sections. 
6. You can only expand the section in 1 level and 2 level, do not expand the section in 3 level or more.
7. Please don't extend the section that is already extended.
8. Just extend one hierarchy level at a time, the subsections you give should not have more than one hierarchy level.
8. The output language must match the language of the user’s query.

** User Query **  
[user query]

** Current Full Report **  
[current survey]

** Available Actions **  
- extend-plan: Expand a section by adding subsections (e.g., section-1 to section-1.1, section-1.2, section-1.3).  
- terminate: No operation.

## Action Format:  
<action> {"name": "expand", "position": "section-x.y.z", "subsections": [{"title": "...", "plan": "..."}, {"title": "...", "plan": "..."}, ...]} </action>  
<action> {"name": "terminate"} </action>

** Output Format **  
<thought> Provide a detailed reasoning process </thought>  
<action> Action (in JSON format) </action>

Please output strictly according to the specified format.
\end{Verbatim}
\end{tcolorbox}
\captionof{figure}{The LLM prompt for the \textit{expand} and \textit{terminate} action.}\label{fig: expand-terminate}

\subsection{User Query Construction}
\label{sec: user_query}

We constructed a dataset of approximately $2000$ user queries with corresponding scoring checklists to support the multi-stage training process. Of these, around $700$ queries are focused on specialized academic survey topics, while the remaining $1300$ address general research reporting topics.

For the academic survey queries, we employed a \textit{reverse question construction} approach: we first selected $700$ surveys from ArXiv and then used a large model to generate user questions based on these articles. The prompts used for this process are available in Figure~\ref{fig: query_gen}. For general research reporting queries, we selected $1300$ real questions.

In addition, we constructed \textbf{a preference checklist} inspired by DeepResearch Bench~\cite{du2025deepresearchbench} for each user query. A large model was used to generate weighted scores for different evaluation aspects based on the existing questions. The final report score for a query is computed as a weighted sum across these aspects. The method for checklist generation is same as DeepResearch Bench.
Out of the $2000$ queries, approximately $1500$ were used to \textbf{construct trajectory data} for SFT and single-step RL, while the remaining $500$ were directly used for end-to-end RL training.

\begin{tcolorbox}[
    breakable, 
    title=Prompt for Search Keywords,
    colback=gray!5!white, 
    colframe=gray!75!black,
    fontupper=\normalsize,
]
\begin{Verbatim}[breaklines=true]
You are a Instruction-writing expert. Below is a Survey title. Your task is to infer the Instruction the user might have.  

Survey Title:  
{title}  

Before crafting the Query, analyze the following:  
1. **Avoid using exact titles.**  
2. Use domain-specific keywords, synonyms.  

First, output your analysis starting with "Thought:", simulating the Survey author’s thought process.  
Then, generate one or more Queries based on the analysis, starting with "Instruction:". DONT OUTPUT EXPLANATIONS AFTER THE Instruction.
\end{Verbatim}
\end{tcolorbox}
\captionof{figure}{The LLM prompt for reverse question (user query) construction.}\label{fig: query_gen}

\subsection{Retrieval Environment Setup}
\label{sec: retrieval_env}

We constructed a local database containing approximately 2.86 million documents to serve as the agent's retrieval environment. This database supports both trajectory data collection and interactive RL training. Among these documents, roughly 2.71 million are abstracts of papers from ArXiv, sourced via Kaggle\footnote{\url{https://www.kaggle.com/api/v1/datasets/download/Cornell-University/arxiv}}
. The remaining 150k documents come from general web pages, for which we employed \textit{Gemini 2.0-Flash} to generate concise summaries while controlling for document length and quality.

To ensure efficient retrieval, we built a vector database. Specifically, all documents were vectorized using the embedding model \texttt{MiniCPM-Embedding-Light} \footnote{\url{https://huggingface.co/openbmb/MiniCPM-Embedding-Light}}
 and indexed with Faiss\footnote{\url{https://github.com/facebookresearch/faiss}}. The pipeline is implemented via \texttt{UltraRAG}\cite{chen2025ultrarag}, an open-source library for constructiong Retrieval-Augmented Generation (RAG) systems.

\subsection{Trajectory Data Construction}
\label{sec: traj_gathering}

Base on the user queries in~\S\ref{sec: user_query} and the retrieval environment in~\S\ref{sec: retrieval_env}, we collected 1,500 actual execution trajectories within our WARP framework.

We chose \textit{Qwen3-235B-A22B-Instruct-2507} as the teacher model, and use the prompts in~\S\ref{sec: prompts}. Despite their scale, current large language models still struggle with high-level decision-making and cannot reliably determine when to stop. To address this, we introduce an \textbf{trajectory pruning} strategy. The gathering process here slightly differs from the standard inference phase of the \ourmethod\ framework: during each \textbf{\textit{Reasoning-Driven Deepening}} stage, we explicitly force the agent to select a position for outline deepening, instead of allowing the model to autonomously decide whether to expand or terminate. This ensures that the agent continuously expands the outline and revises the report until a maximum of $12$ expansions is reached. All the results are scored by the survey-level reward mentioned in~\S\ref{sec: reward_for_report}, and the highest-scoring result is selected as the endpoint of the trajectory.

For each user query, we collect a single high-quality execution trajectory. In total, $1500$ trajectories were obtained corresponding to the $1500$ user queries. Among these, $1200$ trajectories were used as SFT data for cold-start training, while the remaining $300$ were reserved for atomic skill RL.

\subsection{Action Data Distribution}
\label{sec: data_distribution}

For a complete trajectory collected in Section~\ref{sec: traj_gathering}, it typically contains one \textit{initialize} action, one \textit{termination} action, several \textit{expand} actions, and many \textit{search} and \textit{write} actions. These actions correspond to four core agent capabilities: \textit{planning}, \textit{retrieval}, \textit{writing}, and \textit{decision-making}. However, the natural distribution of actions is highly imbalanced with respect to training needs: the more critical and challenging abilities—such as \emph{planning} and \emph{decision-making}—are underrepresented, while easier-to-learn abilities—such as \emph{searching} and \emph{writing}—dominate the trajectories.

To address this issue, we introduce an \textbf{action-level balanced sampling} strategy that increases the sampling probability of more important and difficult actions and decreases that of easier ones, thereby providing more effective supervision for training the agent’s core capabilities.

From the 1,500 collected trajectories, we obtained approximately 100k actions in total. When grouped by user query type, actions from academic review tasks and general report tasks followed an approximate ratio of 3:5. We used about 33k actions for cold-start training (SFT) and about 5k actions for atomic skill RL, with the remaining data discarded.

\subsection{Reward System}

In Reinforcement Learning (RL) training, the design of the reward functions is crucial, often directly impacting training efficiency and stability. 
In this section, we introduce our reward system from two aspects. First, in Section~\ref{sec: reward_for_action}, we introduce our different ability-specific reward functions for four abilities: \textbf{planning, retrieval, writing}, and \textbf{decision-making}, primarily used for single-step RL training. These functions are used for optimizing five actions: \textit{initialize, expand-plan, search, write}, and \textit{terminate}, as shown in Table~\ref{tab: action_ability_mapping}.
Second, in Section~\ref{sec: reward_for_report}, we introduce report-level reward design, which scores the overall quality of the generated report, primarily used for end-to-end RL training.

\begin{table*}[t]
  \small
  \centering
    \caption{The mapping between actions and the four agent abilities.}
  \setlength\tabcolsep{0.7mm}{
  \begin{tabular}{lrl}
    \toprule
    \textbf{Agent Ability} & \textbf{Action} & \textbf{Parameters} \\
    \toprule
    \multirow{2}{*}{planning} 
        & initialize & \{"title": "...", "sections": [\{"title": "...", "plan": "..."\}, \{"title": "...", "plan": "...",\}, ...]\}\\
        & expand & \{"position": "section-x.y.z", "content": "..", "subsections": [\{"title": "..", "plan": ".."\}, ..]\} \\
    
    \multirow{1}{*}{retrieval} 
        & search &  \{"keywords": [keyword-1, keyword-2, ..,]\} \\

    \multirow{1}{*}{writing} 
        & write & \{"position": "section-x.y.z", "title": "...", "content": "..."\}\\

    \multirow{1}{*}{decision-making}
        & terminate &  \{\} \\
    \bottomrule
  \end{tabular}}
  \label{tab: action_ability_mapping}
\end{table*}

\subsubsection{Rewards for Action Optimization}
\label{sec: reward_for_action}

In this section, we introduce our reward design for action optimization, aimed at improving the agent’s four core capabilities.
For LLM as Judgement, the judgment model we use is Qwen2.5-72B-Instruct.

\paragraph{(1) Planning capability.}
\label{sec: reward_for_planning}

Both \textit{Initialize} and \textit{Expand} actions reflect the planning ability, and we calculate the rewards for these two actions by evaluating their generated one-level outlines (with detailed writing plans for each section). We use \textbf{basic properties}, \textbf{holistic quality}, and \textbf{faithfulness} to evaluate the results.

\textbf{Outline Basic Properties} mainly evaluates whether the number of sections in the outline is reasonable, and judges the language consistency. We constrain the number of subsections in a section to between 2 to 7. Meanwhile, we control the language consistency by character statistics.

\textbf{Outline Holistic Quality} is evaluated by LLM for three aspects, following the setting of OmniThink~\citep{xi2025omnithink}. The scoring criteria are in ~\autoref{tab: plan_criteria_rating}, and the prompt~\autoref{fig: plan_eval_prompt}:
\begin{itemize}
    \item \textbf{\textit{Guidance for Content Generation}} Does the outline effectively guide content generation, ensuring comprehensive coverage of the topic?
    \item \textbf{\textit{Hierarchical Clarity}} Does the outline clearly define a hierarchy of topics and subtopics, with a logical, diverse structure that is easy to understand?
    \item \textbf{\textit{Logical Coherence}} Does the outline logically organize topics and subtopics, ensuring a smooth and natural flow of ideas with clear logical transitions?
\end{itemize}
 
\textbf{Outline Faithfulness}~\citep{min2023factscore} verifies whether the content in the plan is real and reliable. 

\paragraph{(2) Retrieval capbility.}
\label{sec: reward_for_retrieval}

\textit{Search} action reflects the retrieval ability, and we calculate the rewards for the action by evaluating its generated search keywords. We use \textbf{recall score} to evaluate them.

\textbf{Recall Score} is calculated by the comparison of the retrieved documents and the golden ones. 

\paragraph{(3) Writing capability.}
\label{sec: reward_for_writing}

The \emph{Write} action reflects the writing ability, and we calculate the rewards for the action by evaluating its generated paragraphs. We use \textbf{basic properties}, \textbf{holistic quality}, \textbf{faithfulness}, and \textbf{citation precision} to evaluate them.

\textbf{Paragraph Basic Properties} mainly evaluates whether the length and the number of citations of a paragraph are reasonable, and judges the language consistency. We constrain the length of a paragraph to between 100 to 2000 tokens. Then, we constrain the citation number of a paragraph to between 0 to 12. Finally, we control the language consistency by character statistics.

\textbf{Paragraph Holistic Quality} is evaluated by LLM for four aspects, according to STORM~\citep{shao2024STORM}, with the prompt~\autoref{fig: content_eval_prompt} and the critria~\autoref{tab: content_criteria_rating} :
\begin{itemize}
    \item \textbf{\textit{Relevance}} How effectively does the report maintain relevance and focus, given the dynamic nature of the discourse?
    \item \textbf{\textit{Coverage}} Does the article provide an in-depth exploration of the topic and have good coverage?
    \item \textbf{\textit{Depth}} How thoroughly does the report explore the initial topic and its related areas, reflecting the dynamic discourse?
    \item \textbf{\textit{Novelty}} Does the report cover novel aspects that relate to the user’s initial intent but are not directly derived from it?
\end{itemize}

\textbf{Paragraph Faithfulness} checks whether the fact claims are consistent with citations.

\textbf{Citation Precision} is calculated by the comparison of the selected citations and the golden ones. We first check the citation hallucination phenomenon, any hallucination will make the score as 0. Then we employ the F1 Score of the citations between the generated content with the golden one.

\paragraph{(4) Decision-making capbility.}
\label{sec: reward_for_planning}

The \emph{terminate} action reflects the agent’s high-level decision-making ability, and its reward is computed by evaluating the correctness of this decision. We use \textbf{accuracy} as the metric: if the agent’s decision matches the reference answer, the reward is $1.0$; otherwise, it is $0.0$.

\subsubsection{Rewards for Pipeline Optimization}
\label{sec: reward_for_report}

Unlike the action-level RL phase, the process-level RL phase no longer has a reference answer to constrain the agent's exploration, and the training data only contains user questions. We will directly evaluate the final result of the entire process (the generated report) from several aspects, instead of scoring the intermediate actions. We hope that in this phase, the agent can explore more freely, generate more diverse results, and ultimately surpass the capabilities of the teacher model. This section will detail our evaluation of the final result from four aspects: comprehensiveness, insight,
instruction-following, and readability. The judgment model we use is \textit{Qwen3-32B}~\citep{yang2025qwen3}.

\section{Experiments Details}
\label{sec: appendix B}

\subsection{Training Details}
\label{app: training_details}

We adopt a three-stage training pipeline consisting of cold start training (SFT), atomic skill RL (single-step RL), and holistic pipeline RL (end-to-end RL). All our experiments are run on 8 A100 GPUs, and the training settings are shown in Table~\ref{tab: exp_settings}.

\paragraph{Cold-Start Training}
For cold-start, we collect approximately 33k action-level samples from 1,200 trajectories. The model is trained using SFT with a learning rate of 1.5e-5 and batch size as 32 for 4 epochs, taking about 2 days to complete.

\paragraph{Atomic Skill RL}
We further perform single-step RL using approximately 5150 action-level samples from 300 trajectories. We set the learning rate to 2.5e-6, batch size to 8, rollout number to 8, and train for 200 optimization steps, taking about 2 days to complete.

\paragraph{Holistic Pipeline RL}
Finally, we conduct end-to-end RL on 500 user queries, optimizing the entire report generation pipeline jointly. The learning rate remains 1e-6, with a batch size of 8, rollout number of 4, and a total of 50 training steps, taking about 4 days to complete.

\begin{table}[t]
  \small
  \centering
    \caption{The training settings for different stages.}
  \setlength\tabcolsep{1.5mm}{
  \begin{tabular}{lrrr}
    \toprule
    \textbf{Parameters} & \textbf{SFT} & \textbf{Single-Step RL} & \textbf{End-to-End RL} \\
    \midrule
    user queries & 1,200 & 300 & 500 \\
    train samples & 33,292 & 5150 & 500 \\
    learning rate & 1.5e-5 & 2.5e-6 & 1e-6 \\
    batch size & 32 & 8 & 8 \\
    rollout number & -- & 8 & 4 \\
    train epochs & 4 & -- & -- \\
    train steps & -- & 200 & 50 \\
    \bottomrule
  \end{tabular}}
  \label{tab: exp_settings}
\end{table}

\subsection{Metrics Details}
\label{app: metrics_details}

We conducted evaluations on three benchmarks: DeepResearch Bench~\cite{du2025deepresearchbench}, DeepConsult~\cite{DeepConsult}, and DeepResearch Gym~\cite{coelho2025deepresearchgym}. 

\paragraph{DeepResearch Bench}
It consists of 100 PhD-level research tasks spanning 22 academic domains. It adopts the RACE and FACT evaluation frameworks. RACE assesses Comprehensiveness, Insight/Depth, Instruction Following, and Readability, while FACT evaluates effective citations per report and citation reliability. We evaluate it by \textit{Gemini-2.5-Pro}.

\paragraph{Deep Consult}
It includes 102 queries from business and consulting scenarios. Evaluation is conducted via pairwise comparisons against an \textit{OpenAI-DeepSearch} baseline, reporting win, tie, and loss rates, together with average quality scores on instruction following, comprehensiveness, completeness, and writing quality. We evaluate it by \textit{o3-mini-2025-01-31}.

\paragraph{DeepResearch Gym}
It is built on the Researchy Questions dataset. Following WebWeaver~\citep{li2025webweaver}, we sample 100 queries from the top 1,000 test queries and evaluate the report quality in six aspects: clarity, depth, balance, breadth, support, and insightfulness. We evaluate it by \textit{GPT-4.1-mini-20250414}.

\begin{table*}[tb]
  \small
  \centering
  \begin{tabular}{p{2cm}p{12cm}}
    \toprule
    \textbf{Rating} & \multicolumn{1}{c}{\textbf{Description}} \\
    
    \midrule
    \multicolumn{2}{c}{\textbf{\textit{Guide}}} \\
    Score 1 & The outline fails to guide content generation, omitting significant aspects of the topic or providing insufficient direction.\\
    Score 2 & The outline provides limited guidance, covering some key areas but lacking depth or completeness in addressing the topic.\\
    Score 3 & The outline provides moderate guidance for content generation, addressing most key areas but leaving some gaps or ambiguities.\\
    Score 4 & The outline effectively guides content generation, covering all significant aspects with clear direction, though minor refinements could enhance comprehensiveness.\\
    Score 5 & The outline is exemplary in guiding content generation, thoroughly addressing all aspects of the topic with clear, detailed direction and no significant gaps.\\ 
    
    \midrule
    \multicolumn{2}{c}{\textbf{\textit{Hierarchical}}} \\
    Score 1 & The outline exhibits no discernible hierarchical structure. Topics and subtopics are jumbled together without logical separation or clear levels, making it nearly impossible to follow or identify any organization.\\
    Score 2 & The outline attempts to establish a hierarchy but fails to maintain logical consistency. Main topics and subtopics are frequently misclassified, and the structure is overly rigid or disjointed. Subtopics may be missing, misplaced, or redundant, making it hard to grasp the intent of the structure.\\
    Score 3 & The outline demonstrates a basic level of logical coherence. Most topics follow a general sequence, but some sections feel forced, with weak or unclear transitions. There are small jumps in logic, causing slight confusion or loss of flow at certain points.\\
    Score 4 & The outline displays a clear, logical, and diverse hierarchical structure. Main topics are distinct, and subtopics are properly nested. While most elements are well-placed, there may be minor redundancies or opportunities to introduce more diverse formats for subtopics. Slight adjustments could achieve better precision and variety in style.\\
    Score 5 & The outline showcases an exceptional, flawless hierarchical structure. Each main topic is distinct, and subtopics are logically nested with absolute clarity and stylistic diversity. The outline demonstrates flexibility in structure and organization, adapting its style where appropriate for the content and logic. No further refinement is necessary.\\
    
    \midrule
    \multicolumn{2}{c}{\textbf{\textit{Coherence}}} \\
    Score 1 & The outline is highly disjointed and incoherent. Topics and subtopics appear in a random, unordered manner, with no logical flow or sense of progression. Major conceptual gaps and illogical jumps are present throughout the structure.\\
    Score 2 & The outline shows some attempt at logical organization, but it contains frequent inconsistencies, abrupt shifts, or logical missteps. Topics and subtopics are misaligned or lack proper transitions, making the reader work hard to follow the structure.\\
    Score 3 & The outline demonstrates a basic level of logical coherence. Most topics follow a general sequence, but some sections feel forced, with weak or unclear transitions. There are small jumps in logic, causing slight confusion or loss of flow at certain points.\\
    Score 4 & The outline exhibits a strong sense of logical flow, with ideas presented in a mostly smooth and connected manner. Transitions between topics and subtopics are clear, but a few minor adjustments could make the flow more seamless or natural. The logic is sound, but room for refinement exists.\\
    Score 5 & The outline achieves exceptional logical coherence. Each topic and subtopic follows a deliberate, thoughtful progression, with clear, natural, and intuitive transitions. The reader experiences a seamless flow of ideas, and no adjustments are required to improve logical consistency or flow.\\

    \bottomrule
  \end{tabular}
  \caption{The plan scoring criteria rating scale 1-5.}
  \label{tab: plan_criteria_rating}
\end{table*}

\begin{table*}[tb]
  \small
  \centering
  \begin{tabular}{p{2cm}p{12cm}}
    \toprule
    \textbf{Rating} & \multicolumn{1}{c}{\textbf{Description}} \\
    
    \midrule
    \multicolumn{2}{c}{\textbf{\textit{Relevance}}} \\
    Score 1 & Very poor focus; discourse diverges significantly from the initial topic and intent with many irrelevant detours.\\
    Score 2 & Poor focus; some relevant information, but many sections diverge from the initial topic.\\
    Score 3 & Moderate focus; mostly stays on topic with occasional digressions that still provide useful information.\\
    Score 4 & Good focus; maintains relevance and focus throughout the discourse with minor divergences that add value.\\
    Score 5 & Excellent focus; consistently relevant and focused discourse, even when exploring divergent but highly pertinent aspects.\\ 
    
    \midrule
    \multicolumn{2}{c}{\textbf{\textit{Coverage}}} \\
    Score 1 & Severely lacking; offers little to no coverage of the topic’s primary aspects, resulting in a very narrow perspective.\\
    Score 2 & Partial coverage; includes some of the topic’s main aspects but misses others, resulting in an incomplete portrayal\\
    Score 3 & Acceptable breadth; covers most main aspects, though it may stray into minor unnecessary details or overlook some relevant points.\\
    Score 4 & Good coverage; achieves broad coverage of the topic, hitting on all major points with minimal extraneous information.\\
    Score 5 & Exemplary in breadth; delivers outstanding coverage, thoroughly detailing all crucial aspects of the topic without including irrelevant information.\\
    
    \midrule
    \multicolumn{2}{c}{\textbf{\textit{Depth}}} \\
    Score 1 & Very superficial; provides only a basic overview with significant gaps in exploration.\\
    Score 2 & Superficial; offers some detail but leaves many important aspects unexplored.\\
    Score 3 & Moderate depth; covers key aspects but may lack detailed exploration in some areas.\\
    Score 4 & Good depth; explores most aspects in detail with minor gaps.\\
    Score 5 & Excellent depth; thoroughly explores all relevant aspects with comprehensive detail, reflecting a deep and dynamic discourse.\\

    \midrule
    \multicolumn{2}{c}{\textbf{\textit{Novelty}}} \\
    score 1 & Lacks novelty; the report strictly follows the user’s initial intent with no additional insights.\\
    score 2 & Minimal novelty; includes few new aspects but they are not significantly related to the initial intent.\\
    score 3 & Moderate novelty; introduces some new aspects that are somewhat related to the initial intent.\\
    score 4 & Good novelty; covers several new aspects that enhance the understanding of the initial intent.\\
    score 5 & Excellent novelty; introduces numerous new aspects that are highly relevant and significantly enrich the initial intent.\\ 
    \bottomrule
  \end{tabular}
  \caption{The content scoring criteria rating scale 1-5.}
  \label{tab: content_criteria_rating}
\end{table*}

\begin{tcolorbox}[
    breakable, 
    title=Prompt for Search Keywords,
    colback=gray!5!white, 
    colframe=gray!75!black,
    fontupper=\normalsize,
]
\begin{Verbatim}[breaklines=true]
###Task Description:
An instruction (might include an Input inside it), a response to evaluate, a reference answer that gets a score of 5, and a score rubric representing a evaluation criteria are given.
1. Identify the major and minor errors in this Response. Write a detailed list of the errors in the response strictly based on the given score rubric, not evaluating in general.
2. After writing the list of errors, write a score that is an integer between 1 and 5. You should refer to the score rubric.
3. The output format should look as follows: "(write the list of errors for criteria) [RESULT] (an integer number between 1 and 5)"
4. Please do not generate any other opening, closing, and explanations.
5. Please be fair, don't hesitate to give a low score like 1 or 2.
6. Note that Major errors refer to actual errors that affects the task severely, may change the meaning of the output, and Minor errors refer to smaller imperfections, and purely subjective opinions about the output. 

###The instruction to evaluate:
{instruction}

###Response to evaluate:
{response}

###Reference Answer (Score 5):
{reference_answer}

###Score Rubrics:
{rubric}

###Feedback: 
\end{Verbatim}
\end{tcolorbox}
\captionof{figure}{The Outline quality reward prompt template.}\label{fig: plan_eval_prompt}

\begin{tcolorbox}[
    breakable, 
    title=Prompt for Search Keywords,
    colback=gray!5!white, 
    colframe=gray!75!black,
    fontupper=\normalsize,
]
\begin{Verbatim}[breaklines=true]
Here is an academic survey about the topic "[TOPIC]":
---
[SURVEY]
---

<instruction>
Please evaluate this survey about the topic "[TOPIC]" based on the criterion provided below, identify the major and minor errors in this survey, and give a score from 1 to 5 according to the score description:
---
Criterion Description: [Criterion Description]
---
Score 1 Description: [Score 1 Description]
Score 2 Description: [Score 2 Description]
Score 3 Description: [Score 3 Description]
Score 4 Description: [Score 4 Description]
Score 5 Description: [Score 5 Description]
---
Note that Major errors refer to actual errors that affect the task severely, may change the meaning of the output, and Minor errors refer to smaller imperfections, and purely subjective opinions about the output. 
There may be multiple errors or no errors in the output. 
After listing the errors, then, please score the survey with 1 to 5. 
Return the score without any other information at the end of the output.
\end{Verbatim}
\end{tcolorbox}
\captionof{figure}{The Content quality reward prompt template.}\label{fig: content_eval_prompt}


\end{document}